\documentclass[runningheads]{llncs}
\usepackage[utf8]{inputenc}
\usepackage{graphicx}
\usepackage{subcaption}
\usepackage{todonotes}
\usepackage{mathtools}
\usepackage{amsfonts}
\usepackage{amssymb}
\usepackage{wrapfig}
\usepackage{times}
\usepackage{url}
\usepackage[margin=17pt]{subcaption}
\usepackage[hidelinks]{hyperref}
\usepackage{color, colortbl}
\usepackage{multirow}
\usepackage{comment}
\usepackage{xcolor}
\usepackage{ulem}

\definecolor{LightCyan}{rgb}{0.9,1,1}
\definecolor{LightGray}{rgb}{0.95,0.95,0.95}

\newcommand{\ME}{\mathcal{E}}

\newcommand{\ML}{\mathcal{L}}
\newcommand{\MK}{\mathcal{K}}
\newcommand{\MY}{\mathcal{Y}}
\newcommand{\BB}{\mathbb{B}}

\begin{document}
\title{Achieving Fairness in Predictive Process Analytics via Adversarial Learning}

%


\author{Massimiliano de Leoni\inst{1} \and Alessandro Padella\inst{1} }
\authorrunning{Massimiliano de Leoni, Alessandro Padella}

\institute{University of Padua, Italy \\  \email{deleoni@math.unipd.it}, \email{alessandro.padella@phd.unipd.it}}

\maketitle

\begin{abstract}
Predictive business process analytics has become important for organizations, offering real-time operational support for their processes. However, these algorithms often perform unfair predictions because they are based on biased variables (e.g., gender or nationality), namely variables embodying discrimination. 
This paper addresses the challenge of integrating a debiasing phase into predictive business process analytics to ensure that predictions are not influenced by biased variables.
Our framework leverages on adversial debiasing is evaluated on four case studies, showing a significant reduction in the contribution of biased variables to the predicted value. The proposed technique is also compared with the state of the art in fairness in process mining, illustrating that our framework allows for a more enhanced level of fairness, while retaining a better prediction quality. 
\end{abstract}
\keywords{Process Mining \and Deep Learning \and Predictive Process Analytics  \and Adversarial Debiasing \and Fairness}
\section{Introduction}\label{sec:introduction}
Predictive process analytics aims to forecast the outcome of running process instances to identify those requiring specific attention, such as instances risking delays, excessive costs, or unsatisfactory outcomes. By proactively predicting process behavior and outcomes, predictive process analytics enables timely intervention and informed decision-making. Over the years, numerous approaches have been proposed in the literature to address the challenges associated with predictive process analytics (cf. Section \ref{sec:related_works}).

Predictive process analytics naturally needs to rely onto the characteristics of the process being monitored, and performs predictions on their basis. Being that said, this analytics become a problem when predictions are unfair because they are based on characteristics that discriminate in a form that is unacceptable from a legal and/or ethical point of view. For instance, in a loan-application process at a financial institute, one cannot build on the applicant's gender to predict the outcome, namely whether or not the loan is granted. 
Pohl et al.\ indicate monitoring, detecting and rectifying biased patterns to be the most significant challenge in Discrimination-Aware Process Mining~\cite{10.1007/978-3-031-27815-0_8}.

Process characteristics are hereafter modelled as process variables. In accordance with the literature terminology~\cite{10.1145/3278721.3278779}, we use the term \textit{protected variable} to indicate the variables on which prediction cannot be based.
The choice of the set of variables to protect depends on the specific process, and thus needs to be made by the process analysts/stakeholders. Note how simply removing the protected variables from the datasets would not be effective, because the bias would be simply ``hidden under the carpet'', as it would be possibly just transferred to other variables that are strongly correlated to the protected variables~\cite{10.1007/s10115-017-1116-3}. 

While several researchers acknowledge the importance of ensuring fairness in process predictive analytics, very little research has been carried out on this topic: the state of the art only proposes simple Machine-Learning models, such as decision trees, that are unable to guarantee accurate predictions and high levels of fairness (cf.\ discussion in Section~\ref{sec:related_works}). 
This paper proposes a framework based on  \emph{adversarial debiasing}, which aims to mitigate bias related to protected variables within the predictive models. In a nutshell, the proposed framework is based on the idea to train the model to predict the process' outcome values while constraining a jointly-trained adversary from accurately predicting the protected variables, reducing bias in its learned representations.

Compared with the current literature in fairness for process' predictive analytics, adversial debiasing aims at more accurate predictions through prediction models that also guarantee higher fairness. However, existing research on adversial debiasing has not focused on process predictive analytics  and, more generally, to time series, and cannot be trivially applied in this setting (cf.~Section~\ref{sec:related_works}).

Experiments have been conducted on four case studies to forecast the process-instance total time and whether or not certain activities are eventually going to occur. Protected variables accounted for resources, organization countries, gender, citizenship and spoken languages. 
The results show that our framework ensures fairness with respect to the chosen protected variables, while the accuracy of the predictive models remain high, also in comparison with the results for research works in literature that tackle the problem to ensure fairness in predictive process analytics. The experimental results also highlight that the influence is also reduced for those process' variables that are strongly correlated with the protected variables. This illustrates how indeed simply removing the protected variables and values would just transfer the unfairness to the correlated variables. 

Section~\ref{sec:related_works} presents related work in fairness in both Machine Learning and and Process Mining. Section~\ref{sec:preliminaries} discusses the necessary background. Section~\ref{sec:adv_framework} introduces the framework embodying adversial debiasing to achieve higher fairness in process predictive analytics. Section~\ref{sec:evaluation_framework} illustrates the evaluation methods and results. Finally, Section~\ref{sec:conclusion} summarizes the contributions and the experimental results, also highlighting the limitations and delineating the future-work directions.

\section{Related Works}\label{sec:related_works}
The existing literature related to debiasing in process mining is relatively limited: Mannhardt in~\cite{Mannhardt2022} offers a comprehensive examination of Responsible Process Mining, where algorithmic fairness is identified as one of the four topics in which Responsible process mining is divided. As mentioned in Section~\ref{sec:introduction}, Qafari et al. in~\cite{10.1007/978-3-030-33246-4_11} is the only framework for fairness in process mining prescriptive analytics. The proposed approach operates through a process of identifying and rectifying mislabeled instances within the training data, achieved via an iterative traversal of the decision tree predictor. During this traversal, the labels of the leaf nodes are examined, and if a significant number of erroneous labels are detected within the fairness context, the labels of that particular node are subsequently adjusted to align with the majority class.

Several Deep learning techniques have been proposed in the literature to address bias within AI frameworks. The idea of adversarial debiasing has been proposed by Zhang et al.~\cite{10.1145/3278721.3278779} and by Beutel et al.~\cite{BIAS_2}.
However, both research works have not focused on debiasing predictions of processes or, more generally speaking, prediction of temporal series: they rather focused on, e.g., word embedding~\cite{10.1145/3278721.3278779}, or predicting the monetary income based on the characteristics of the people's CV~\cite{BIAS_2}.
Their proposal could not thus be directly applied, and the effectiveness for process predictive was not trivially valid. 


Alternative approaches exist in literature for debiasing in Machine and Deep Learning. Li et al.\ in~\cite{10.1145/3485447.3511958} leverages on a Generative Adversarial Networks based learning algorithm, named FairGAN, that dynamically generate appropriate data points to fairly train the predictive model. 
The data points in Predictive Analytics are the log's events, and their artificial generation is hard because they cannot be independently generated, because events are naturally dependent on other events (cf.\ events of the same event-log trace). Therefore, we found this framework very difficult to be used in our context. Yang et al.\ propose a framework for mitigating algorithmic bias in clinical Machine learning using Deep Reinforcement Learning(deep RL)~\cite{reinf_learning}. The framework is based on the idea of using a RL agent to learn a policy that minimizes bias in a Machine learning model. Nevertheless, the survey reported in~\cite{10.1145/3616865} indicates that models grounded in RL consistently exhibit inferior performance compared to those based on adversarial training.

\section{Preliminaries}\label{sec:preliminaries}
The starting point for a prediction system is an \textit{event log}. 
An event log is a multiset of \emph{traces}. Each trace describes the life-cycle of a particular \emph{process instance} (i.e., a \emph{case}) in terms of the \emph{activities} executed and the process \emph{attributes} that are manipulated.


\begin{definition}[Events]\label{def:event}
Let $\mathcal{A}$ be the set of process activities and let $\mathcal{V}$ be the set of process attributes. Let $\mathcal{W}_\mathcal{V}$ be a function that assigns a domain $\mathcal{W}_\mathcal{V}(x)$ to each process attribute $x\in\mathcal{V}$. Let $\overline{\mathcal{W}} = \cup_{x\in\mathcal{V}}\mathcal{W}_\mathcal{V}(x)$. An event is a pair $(a,v)\in\mathcal{A}\times(\mathcal{V}\not\to\overline{\mathcal{W}})$ where $a$ is the event activity and $v$ is a partial function assigning values to process attributes with $v(x)\in\mathcal{W}_{\mathcal{V}}(x)$.
\end{definition}
Note that the same event can potentially occur in different traces, namely attributes are given the same assignment in different traces. 
This means that potentially the entire same trace can appear multiple times. This motivates why an event log is to be defined as a multiset of traces.

\begin{definition}[Traces \& Event Logs]
Let $\ME$ be the universe of events.
A trace $\sigma$ is a sequence of events, i.e.\ $\sigma \in \ME^*$.
An event-log $\ML$ is a multiset of traces, i.e.\ $\ML \in \BB(\ME^*)$\footnote{Given a set $A$, $\BB(A)$ indicates the set of all multisets with the elements in $A$.}.
\end{definition} 

Given an event $e=(a,v)$, the remainder uses the following shortcuts: $activity(e)=a$, $variables(e)=v$ and, given a trace $\sigma=\left\langle e_{1}, \ldots, e_{n}\right\rangle$, $prefix(\sigma)$ denotes the set of all prefixes of $prefix(\sigma)=\sigma$, including $\sigma$: $\left\{\langle\rangle,\left\langle e_{1}\right\rangle,\left\langle e_{1}, e_{2}\right\rangle, \ldots,\left\langle e_{1}, \ldots, e_{n}\right\rangle\right\}$.

Predictive Process Analytics aims to address the prediction problem:
\begin{definition}[The Process Prediction Problem]
Let $\ME$ be the universe of events.
Let $\MK: \ME^* \rightarrow \mathcal{O}$ be an outcome function that, given a trace $\sigma \in \ME^*$, measures the actual $\sigma$'s outcome $\MK(\sigma)$, which is a value in a outcome domain $\mathcal{O}$.
Let $\sigma'=\langle e_1,\ldots,e_k\rangle$ be the trace of a running case, which eventually will complete as $\sigma_T=\langle e_1,\ldots,e_k, e_{k+1}\ldots,e_n \rangle$. The prediction problem aims to forecast the value of $\MK(\sigma_T)$, after observing the prefix $\sigma'$. 
 \label{def:prediction_problem}
 \end{definition}
To tackle the process prediction problem for an outcome function $\MK: \ME^* \rightarrow \mathcal{O}$, we need to build a \textbf{process prediction oracle} $\Psi_\MK: \ME^* \rightarrow \mathcal{O}$ such that, given a running trace $\sigma'$ eventually completing in $\sigma_T$, $\Psi_\MK(\sigma')$ is a good predictor of $\MK(\sigma_T)$.

The literature proposes several Machine- and Deep-Learning techniques~\cite{10.1007/978-3-031-07472-1_18} for this aim, where Long Short-Term Memory (LSTM) networks have shown excellent predictive power (see, e.g.,~\cite{LSTM_camargo,DBLP:conf/caise/TaxVRD17}.

In the repertoire of neural networks, we opted for fully connected neural networks (FCNNs)~\cite{fcnn}, which are faster to train than LSTM networks but provide similar accuracy results (see our comparison reported in Section~\ref{subsec:lstm_times}).
Also known as Feed-Forward Neural Networks, FCNNs are characterized by having every node in one layer connected to every node in the next layer. This means that every node in one layer receives input from every node in the previous layer and produces an output that is sent to every node in the next layer. FCNNs show impressive computational power, 
excelling at capturing non-linear relationships in data. This makes them suitable for data that exhibit intricate patterns and interactions.

Note how our framework does not support predictive models that are not based on  neural networks, such as random forests, support vector machines, or regression techniques~\cite{10.1007/978-3-031-07472-1_18}. The main reason is that, our framework alters the nodes of the predictive models to add connections to nodes of a second neural network, namely the adversarial network (cf.\ Section~\ref{sec:adv_framework}).

\subsection{Predictive Process Analytics via Fully Connected Neural Networks}\label{sucsec:preprocessing}
The training of FCNN models falls into the problem of supervised learning, which aims to estimate a Machine-Learning (ML) function $\Phi:X_1\times\ldots\times X_n\rightarrow \MY$ where $\MY$ is the domain of variable to predict (a.k.a.\ dependent variable), and $X_1\ldots X_n$ are the domains of some independent variables $V_1, \ldots, V_n$, respectively.

To tackle the prediction problem for an outcome function $\MK: \ME^* \rightarrow \mathcal{O}$,  $\MY=\mathcal{O}$. The values of the independent variables are obtained from the event-log traces: each  trace is encoded into a vector element of $X_1\times\ldots\times X_n$, through a \textbf{trace-to-instance encoding function} $\rho_\ML: \mathcal{E}^* \rightarrow X_1\times\ldots\times X_n$. Note that the process prediction oracle is thus implemented as $\Psi_\MK(\sigma)=\Phi(\rho_\ML(\sigma))$.

Several alternatives exist to define encoding function $\rho_\ML$~\cite{barbon2023trace}: here, we use the most widespread (cf.\ survey by M\'{a}rquest-Chamorro et al.~\cite{Marquez-Chamorro18}). The encoded vectors are in the domain $X_{v_1} \times \ldots \times X_{v_m} \times \ldots X_{a_1} \times X_{a_p}$ such that there is a variable $X_{v_i}$ for each process attribute $v_i \in \mathcal{V}$ and there is a variable $X_{a_i}$ for each process activity $a_i \in \mathcal{A}$. 
Given a trace $\sigma=\langle e_1, \ldots, e_n \rangle$, the vector $(x_{v_1}, \ldots, x_{v_m}, \ldots x_{a_1}, \ldots, x_{a_p}) = \rho_\ML(\sigma)$ is such that \textit{(i)} $x_{a_i}$ takes on a value equal to the number of events $e \in \sigma$ with $activity(e)=a_i$, and \textit{(ii)} $x_{v_i}$ is the latest value assigned to variable $v_i$ by $\sigma$ (i.e.\ there is an index $1 < j \leq n$ with $variable(e_j)(v_i)=x_{v_i}$ and, for all $j < k \leq n$, $v_i$ is not in the domain of $variable(e_k)$).

The prediction model for any ML function $\Phi:X_1\times\ldots\times X_n\rightarrow \MY$ is trained via a multiset $\mathcal{D}$ of instances belonging to $(X_1\times\ldots\times X_n \times \MY)$. 
For predictive process analytics, $\mathcal{D}$ is created from a training
event log $\ML\in\mathbb{B}(\mathcal{E}^*)$ as follows: each prefix $\sigma^{p}$ of each trace $\sigma \in \ML$ generates one distinct element in $\mathcal{D}$ consisting of a pair $(\vec x,y) \in ((X_1\times\ldots\times X_n) \times \MY)$ where $\vec x=\rho_\ML(\sigma^{p})$ and $y=\mathcal{O}(\sigma)$.


\begin{figure}[t!]
    \centering
    \includegraphics[width=1\linewidth]{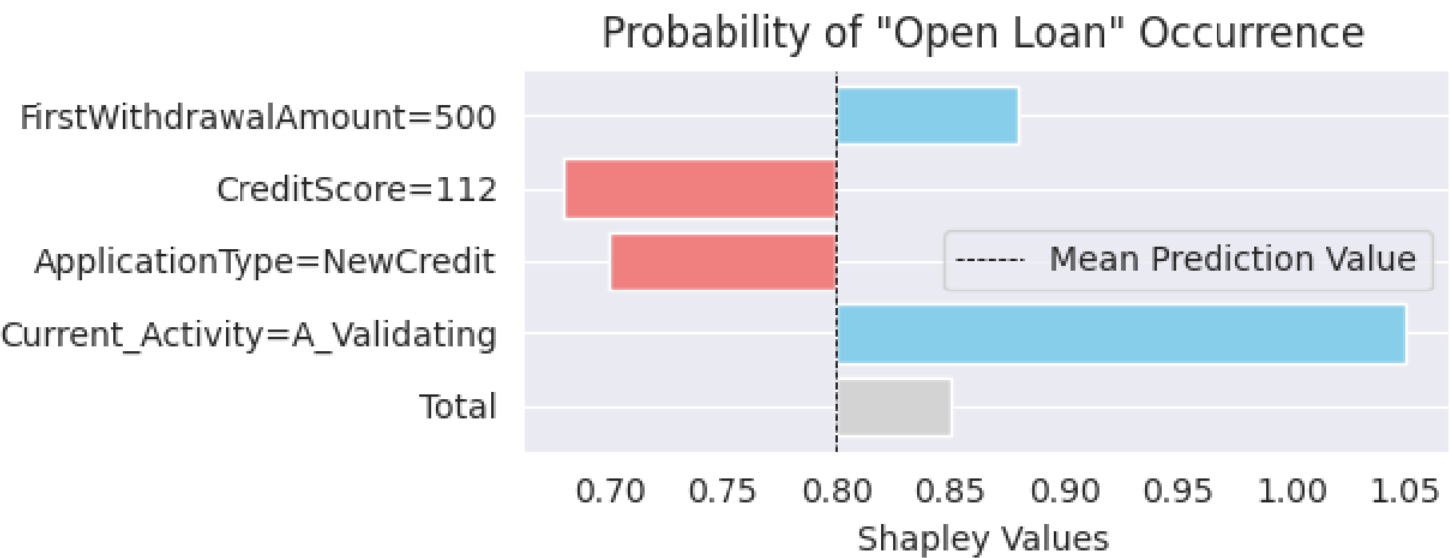}
    \caption{Example of Shapley Values in the prediction of the probability of occurrence for the activity ``Open Loan" in a loan application process. The y-axis denotes the variable names assuming a certain value, while the x-axis represents the probability values. Shapley Values indicate deviations from the mean prediction value, that is 0.80.}
    \label{fig:shap}
\end{figure} 

\subsection{Assessment of the Variable Influence on Predictions}\label{subsec:shap_intro}
The evaluation leverages on computing the influence of each variable $V_1, \ldots, V_n$ on the predictions returned by the ML function $\Phi:X_1\times\ldots\times X_n\rightarrow \MY$ where $X_1, \ldots, X_n$ are again the domains of the variables $V_1, \ldots, V_n$, respectively.
In particular, we will assess the influence of the protected variables on the predictions and show how this is reduced after employing our debiasing technique. 

For evaluating the influence of a variable, we use the widely adopted technique of Shapley Values~\cite{shap,DBLP:journals/corr/abs-2105-07168}. Shapley values are computed for each variable for each instance separately. Let  $(x_1,\ldots,x_n)$ be an instance defined over variables $V_1,\ldots,V_n$, which is predicted to return $y=\Phi((x_1,\ldots,x_n))$. The Shapley value of any variable $V_i$ is the contribution of $V_i=x_i$ to the prediction, where the contribution measures how much $V_i=x_i$ makes prediction $y$ deviate from the mean prediction value. It follows that larger deviations imply higher influence on the prediction.

Consider the example presented in Figure~\ref{fig:shap}, focusing on the prediction of the probability of ``Open Loan'' occurrence within a loan application process. Initially set at 0.8, the mean prediction value for predictions undergoes adjustments based on individual variable contributions. For instance, the  \textit{CreditScore} variable taking a value of 112, diminishes the probability of loan approval by 0.11. Conversely, the state in which the procedure has been validated by the relevant authorities positively influences the probability by 0.25. The cumulative effect of all these contributions results in a deviation of 0.05 from the initial mean prediction value, positively impacting the prediction.

The concept can be generalized to any variable $V_i$, irrespectively of the value $x_i$ taken on by $V_i$, through a support set $T \in \mathbb{B}(X_1\times\ldots\times X_n)$: for each vector \linebreak $(x_1,\ldots,x_n)~\in~T$, we consider the Shapley value $V_i=x_i$ and, then, we compute the mean value for all vectors in $T$.

\section{An Adversarial Debiasing Framework for Predictive Process Analytics}\label{sec:adv_framework}
The overall objective of this paper is to build a process prediction function $\Psi_\MK$ whose output values are not influenced by the chosen \textbf{protected variables}.

The determination of the protected variable depends on the the specific case study under consideration (e.g., the gender or nationality of a loan applicant). It is crucial to note that certain variables may be designated as protected in one case study but not in another. For instance, the variable ``Gender'' might be designated as a protected variable in the context of a loan application process, but it may not hold the same status in the process of hospital discharge. By carefully selecting the protected variables, we aim to ensure that the predictions do not enforce a discrimination that is not ethically and/or morally acceptable.

\begin{figure}[t!]
    \centering
    \includegraphics[width=1\linewidth]{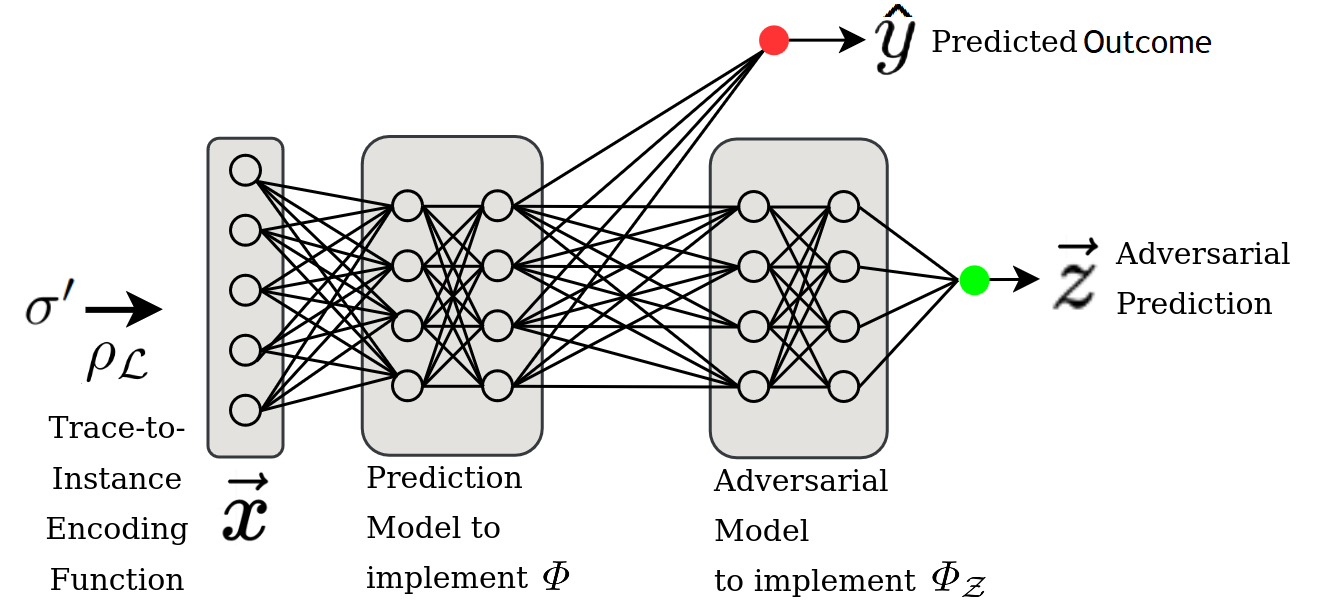}
    \caption{The figure provides an overview of our debiasing framework for process' predictive analytics. Vector $\vec x$ is the encoding of the sequence of events of a running case. Two FCNN models are used within our framework, where $\hat y$ is the predicted outcome value, and $\vec z$ is the forecast of the values of the protected variables in $\vec x$, given the output of the last layer of the FCNN implementing $\Phi$. The overall model aims to accurately predict $\hat y$ while scoring poor to predict $\vec z$. The dots indicate the encoding layers to generate $\hat y$ and $\vec z$.}
    \label{fig:framework}
\end{figure}

The framework is visually depicted in Figure~\ref{fig:framework} where the core component is the prediction model that implements the oracle function  $\Psi_{\MK}$, capable to 
of forecasting the outcome of a running trace.
Leveraging on neural networks, $\Psi_\MK$ is obtained through the composition of the trace-to-instance encoding function $\rho_L$ and an ML function \mbox{$\Phi: X_1 \times\ldots\times X_n \rightarrow \mathcal{O}$}, namely for any trace $\sigma$ $\Psi_\MK(\sigma)=\Phi(\rho_L(\sigma))$. The most left gray box in  Figure~\ref{fig:framework} is the encoder $\rho_L$, which converts the trace into a vector. The second gray box from left depicts the FCNN that implements $\Phi$, along with the decoder represented through the red dot. 

Looking from the right in Figure~\ref{fig:framework}, the first gray box depicts the adversarial FCNN, which tackle the debiasing problem to ensure fairness. In particular, let $\overline V=\{\overline{V_1},\ldots, \overline{V_p}\}$ \allowbreak $\subseteq\{V_1,\ldots,V_n\}$ be the set of the protected variables, which are defined over the domains $Z=\overline{X_1},\ldots, \overline{X_p}$, respectively.
Let $N_1, \ldots, N_q$ are the domains of the output of the $q$ nodes that constitute the last layer of the FCNN implementing $\Phi$. The adversarial FCNN implements a function $\Phi_Z: N_1 \times N_n \rightarrow Z$, which aims to predict the values of the protected variables, using the output of the last layer as input. 

In accordance with the literature on adversarial debiasing~\cite{10.1145/3278721.3278779}, if the neural network that implements $\Phi$~-~in our case a FCNN~-~ does not build the prediction on the protected variables, then the adversarial network that implements $\Phi_Z$~-~in our case another FCNN~-~is unable to predict the protected-variables values from the output of the network implementing $\Phi$. 

More formally, let $\hat y=\Phi(\vec x)$ be the predicted value for the running trace $\sigma'$ that has been encoded $\vec x = \rho_L(\sigma')$. Let $\sigma$ be the real completion of $\sigma'$ (i.e.\ $\sigma'$ is a prefix of $\sigma$), with the real outcome $y=\MK(\sigma)$. Let $\vec z=\Phi_Z(\vec n)$ be the vector of the values predicted for the protected variables, on the basis of the vector $\vec n$ of the output of the last layer of the neural network that implements $\Phi$. The two neural networks are trained so as to minimize the overall loss function:
\begin{equation}\label{eq:loss}
    L_{\overline V}(\hat y, y, \vec x, \vec z)= \Delta(\hat y,y) - \Delta(\vec z,\pi_{\overline{V}}(\vec x)).
\end{equation}
where $\Delta$ indicates the normalized difference between two vectors (or two values), and $\pi_{\overline{V}}(\vec x)$ is the projection of $\vec x$ over $\overline V$, namely retaining the dimensions of $\vec x$ for the protected variables. The normalization in $\Delta(\hat y,y)$ is performed by dividing by the largest outcome value $y=\MK(\sigma)$ for all traces $\sigma$ in the training event log. The normalization in $\Delta(\vec z,\pi_{\overline{V}}(\vec x))$ is achieved by dividing by the largest vector $\pi_{\overline{V}}(\rho_L(\sigma))$ for all traces $\sigma$ in the training event log.


Minimizing Equation~\ref{eq:loss} implies in essence that prediction accuracy is kept reasonably high while the influence of protected variables is minimized. 

The whole framework has been implemented through the training of two FCNNs on a stochastic-gradient-descent based algorithm. The inplementation is in Python and available at \url{https://anonymous.4open.science/r/Fairness-D70B}, \linebreak leveraging on the 
\textit{PyTorch} package for FCNN's training and \textit{fairlearn} for other debiasing utilities.



\section{Evaluation}\label{sec:evaluation_framework}
The evaluation focuses on evaluating how our framework mitigates the influence of protected variables while still ensuring a good quality.



The framework evaluation was carrying out, as previously mentioned, by training two FCNNs that implement functions $\Phi$ and $\Phi_Z$. In particular, we carried out a grid search to tune the hyper-parameters related to the learning rate, layers shape, epochs, and weight decay, so as to prevent over- and under-fitting problems. 

Our debiasing framework was evaluated on four case studies, aiming to assess \textit{(i)} the mitigated influence of the protected variables on the prediction, and \textit{(ii)} the extent of the reduction of the prediction accuracy when our framework was employed. Note that a reduction in accuracy is expected when addressing the fairness problem: if the protected variables have some good predictive power, their exclusion has a natural negative impact on the ML-model accuracy.
The baseline of comparison is with the only existing framework by Qafari et al.~\cite{10.1007/978-3-030-33246-4_11}.

The remainder of this section is organized as follows. Section~\ref{subsec:case_studies} introduces the case studies and the train-test splitting of event logs, while Section~\ref{subsec:protv} discusses the choice of the protected variables. Section~\ref{subsec:shap_e_odds} reports on the metrics used for the evaluation, while Section~\ref{subsec:results} details and analyses the results. In Section~\ref{subsec:lstm_times}, training times and accuracies for both LSTM and FCNN models are presented, motivating the FCNN choice.

\subsection{Introduction to Use Cases and Event-log Datasets}\label{subsec:case_studies}
Our technique was assessed through three process for which we have identified four case studies. The first and the second case study are from Volvo Belgium and refer to a process that focuses on an incident and problem management system called \texttt{VINST}.\footnote{\nolinkurl{https://data.4tu.nl/articles/dataset/BPI_Challenge_2013_incidents/12693914}} Executions of this process are recorded in an event log with 7,456 completed traces and 64,975 events. The process consists of 13 different activities that can be accomplished by 649 resources. In the first case study, our aim is to predict the \textbf{total time} of an execution that is running, while in the second our aim is to predict \textbf{whether or not the activity \textit{Awaiting Assignment} will occur} in the future for the same process instance. When that activity occurs, it means that the procedure has been assigned to the wrong division of the Volvo system, negatively affecting in terms of time and costs.

The third case study refers to the \textit{Hiring} process provided by Pohl et al.\ in~\cite{pohl}. It deals with a multi-faced requirement procedure with diverse application pathways. Executions of this process are recorded in event log containing 10,000 completed traces and 69,528 events. There are 12 different activities that can be accomplished by 8 different resources. For this case study we aim to predict the \textbf{total time} a running execution. 

The last case study is based on the \textit{Hospital} process discussed by Pohl et al.~\cite{pohl}. It depicts a hospital treatment that starts with registration at an Emergency Room or Family Department and advances through stages of examination, diagnosis, and treatment. Executions of this process are recorded in 10,000 completed traces and 69,528 events. There are 10 different activities that can be accomplished by 7 different resources. For this case study our aim is to predict \textbf{whether or not the activity \textit{Treatment unsuccessful} will occur} in the future. 

Note that, in the event logs taken from Pohl et al.\ in~\cite{pohl}, we removed the variables ``activity" and ``time" because they were identical to respectively ``concept:name" and ``time:timestamp", along with the variable ``@@index", which simply is an absolute, progressive numbering of the log events, which, observing no concept drifts, has no discriminating power.

For each case study, the available process log has been split into 70\% of the traces that were used for training the prediction and adversarial models and 30\% for testing. The splitting is based on time, considering the timestamp of the first trace's event: the earliest 70\% of the traces are part of the training log, and the latest 30\% is part of the test log.  
The last column in Table~\ref{tab:protv} reports the number of traces in each log.

\subsection{Selection of Protected Variables}\label{subsec:protv}
Each case study clearly uses different protected variables, since their choice depend on process and is also related to specific fairness-preserving considerations. 
Table~\ref{tab:protv} summarizes the choices for the four case studies.

For the \textit{VINST} process, we opted to protect variable \textit{resource country} when the predicted outcome is the total process-instance execution time: that variable encodes the country of residence of the resource in the support team working on the case: ensuring fairness for this case study means that the prediction of the total case is not biased by the residence country of the resource that manages the request. For the same process, we also used \textit{organization country} as alternative protected variable when the predicted outcome is whether or not the process-instance execution has observed \textit{Awaiting Assignment}, an undesired activity linked to a downtime. The choice of a different protected variable is to illustrate that our technique also works when the protected variable or variables are altered. 
Variable \textit{organization country} stores the location that takes ownership of the support team and the objective: ensuring fairness with respect to this variable means that the prediction whether or not the undesired activity occurrence is executed is not influenced by the country where the request is made.

For the \textit{Hiring} and \textit{Hospital} processes, we follow the indication given by Pohl et al.~\cite{pohl}. For the former we aim to protect variables \textit{Gender} and \textit{Religious} (i.e., whether or not the patient is religious); for the \textit{Hospital} process, we protect two boolean variables: \textit{Citizen} (i.e., whether or not the patient is German) and \textit{german\_speaking} (i.e., whether the patient does or does not speak German).

\begin{table}[t!]
    \centering
    \resizebox{\columnwidth}{!}
{
    \begin{tabular}{|c|c|c|l|} \hline 
 \textbf{Process}& \textbf{Predicted Outcome} & \textbf{Protected Variables}  &\textbf{Train/Test Cases}\\ \hline 
 VINST& Total Time&\textit{Resource Country} &5,219/2,237\\ \hline 
 VINST& Occurrence of \textit{Awaiting Assignement} &\textit{Organization Country} &5,219/2,237\\ \hline 
 Hiring& Total Time& \textit{Gender} and \textit{Religious} &7,000/3,000\\ \hline 
 Hospital& Occurrence of \textit{Treatment Unsuccessful} & \textit{Citizen} and \textit{german\_speaking}  & 7,000/3,000\\ \hline
    \end{tabular}
}
\caption{Summary of different case studies with the chosen outcome: the variables protected variables in the second, and the protected variables in the third. The train/test cases column indicates the sample sizes for training and testing datasets in each case study.}
\label{tab:protv}    
\end{table}


\subsection{Evaluation Metrics}\label{subsec:shap_e_odds}
The evaluation's goal is twofold, as indicated at the beginning of the section, and aims to assess
the mitigation influence of the protected variables on the prediction, and the extent of the reduction of the prediction accuracy when our framework was employed.

For the first and third case studies in which we aim to predict the total time of running traces, i.e. a regression problem, the results are provided in terms of \textbf{Absolute Percentage Accuracy} (APA) , which is defined as $100\%$ minus Mean Absolute Percentage Error, between the actual value and the predicted one. For the second and fourth case studies, we aim to test the accuracy prediction on the occurrence for the activities \textit{Awaiting Assignment} and \textit{Treatment unsuccessful}, respectively. This is a classification problem: hence, we choose \textbf{F-score} for assessing the accuracy of our predictions. 

To assess the reduction in the influence of protected variables, we employ the theory of Shapley values (cf.\ Section~\ref{subsec:shap_intro}), computing them both when our framework is employed and when it is not: our framework is expected to reduce the absolute Shapley value, which corresponds to a lower influence. For classification problems, we also assess an enhanced fairness through the analysis of the false positive rate (FPR) and true positive rate (TPR), and the verification of the \textbf{Equalized Odds} criterion~\cite{10.1145/3278721.3278779}: this criterion states that, if we group the samples in the test set by the values of the protected variables, the FPR and TPR should be somewhat similar in all groups. The rationale behind this criterion is that, splitting the test-set samples based on the values of the protected variables, one obtains groups that are statistically equated, including for false and true positive rates, if the model's prediction are not based on the protected variables.

\begin{table}[t!]
    \centering
    \resizebox{\columnwidth}{!}
{
    \begin{tabular}{|c|c|c|c|c|c|} \hline 
        \textbf{Process} & \textbf{Outcome} &  \textbf{Methodology} & \textbf{Without} & \textbf{With} & $\Delta$\\ \hline 
        \multirow{2}{*}{VINST} & \multirow{2}{*}{Total Time} & Qafari et al.~\cite{10.1007/978-3-030-33246-4_11} & 69\% & 60\% &  9\%\\ 
        \cline{3-6}
        & & Our Framework & 78\% & 74\% &  4\%\\ \hline 
        \multirow{2}{*}{VINST} &  Occurrence of & Qafari et al.~\cite{10.1007/978-3-030-33246-4_11} & 0.71 & 0.59 &  0.12\\ 
        \cline{3-6}
        & \textit{Awaiting Assignment} & Our Framework & 0.80 & 0.72 &  0.08\\ \hline 
        \multirow{2}{*}{Hiring} & \multirow{2}{*}{Total Time} & Qafari et al.~\cite{10.1007/978-3-030-33246-4_11} & 79.9\% & 70.02\% &  9.88\%\\ 
        \cline{3-6} 
        & & Our Framework & 83.6\% & 81.1\% &  2.5\%\\ \hline 
        \multirow{2}{*}{Hospital} & Occurrence of & Qafari et al.~\cite{10.1007/978-3-030-33246-4_11} & 0.69 & 0.58 &  0.11\\ 
        \cline{3-6}
        &  \textit{Treatment Unsuccessful} & Our Framework & 0.78 & 0.76 &  0.02\\ \hline
    \end{tabular}
}
\caption{Results achieved by our framework and by Qafari et al.~\cite{10.1007/978-3-030-33246-4_11}, in terms of accuracy. The first and third rows provide results in terms of Absolute Percentage Accuracy. The second and the fourth rows provide results in terms of F-score. The columns \textit{without} and \textit{with} show the results when the framework is not or is used, respectively; column $\Delta$ highlights their difference.}
    \label{tab:accuracy_scores}
\end{table}

\subsection{Evaluation Results}\label{subsec:results}
Table~\ref{tab:accuracy_scores} illustrates the results in terms of accuracy for the processes, logs and predicted outcomes introduced in Section~\ref{subsec:case_studies}. The results are based on a test set that is are constructed as discussed in Section~\ref{subsec:case_studies}, and they refer to the work proposed in this paper, which is then compared with the results that Qafari et al.~\cite{10.1007/978-3-030-33246-4_11} can achieve, which is considered as baseline. Columns \textit{without} and \textit{within} report on the results when the corresponding techniques doesn't or does aim at achieving fairness, respectively. Column $\Delta$ highlights the reduction of accuracy when the techniques aims at fairness. \textit{Our framework consistently obtains higher accuracy for all case studies, if compared with Qafari et al.~\cite{10.1007/978-3-030-33246-4_11}, and also the accuracy reduction is significantly more limited}.

\begin{table}[t!]
    \centering
    \resizebox{\columnwidth}{!}
{
    \begin{tabular}{|c|c|c|c|c|c|} \hline 
         \textbf{Process} & \textbf{Outcome} &  \textbf{Protected Variable}& \textbf{Without}&    \textbf{With} & \textbf{Ratio}\\ \hline 
         VINST&  Total Time&Resource country&  112h& 9h & 8\%\\ \hline 
         \multirow{2}{*}{VINST} &  Occurrence of &  Organization &  \multirow{2}{*}{1.8}&\multirow{2}{*}{0.03} & \multirow{2}{*}{1\%} \\ 
         & \textit{Awaiting Assignment}  & country &&&\\
         \hline 
         \multirow{2}{*}{Hiring}&  \multirow{2}{*}{Total Time}&  Gender&  -463min&-156min & 20\% \\ \cline{3-5} 
         &  &  Religious&  -447min&-12min & 3\% \\ \hline 
         \multirow{2}{*}{Hospital}&  Occurrence of &  Citizen&  0.25&0.04 & 16\%\\ \cline{3-5}
         & \textit{Treatment Unsuccessful}  &  german\_speaking&  0.17&0.06 & 35\%\\ 
         \hline
    \end{tabular}
}
\caption{Differences in Shapley Values of protected variables with and without the debiasing framework, for the four case studies. The columns \textit{without} and \textit{with} show the results when the framework is not or is used, respectively, where column $Ratio$ highlights the ratio between the respective Shapley value without and with using the framework. The significant low ratio, especially in the first two case studies and in our protected variable of the the third, illustrates the high debiasing effectiveness of our framework.}
    \label{tab:shap_drop}
\end{table}

\begin{figure}[t!]
    \centering
    \includegraphics[width=\linewidth]{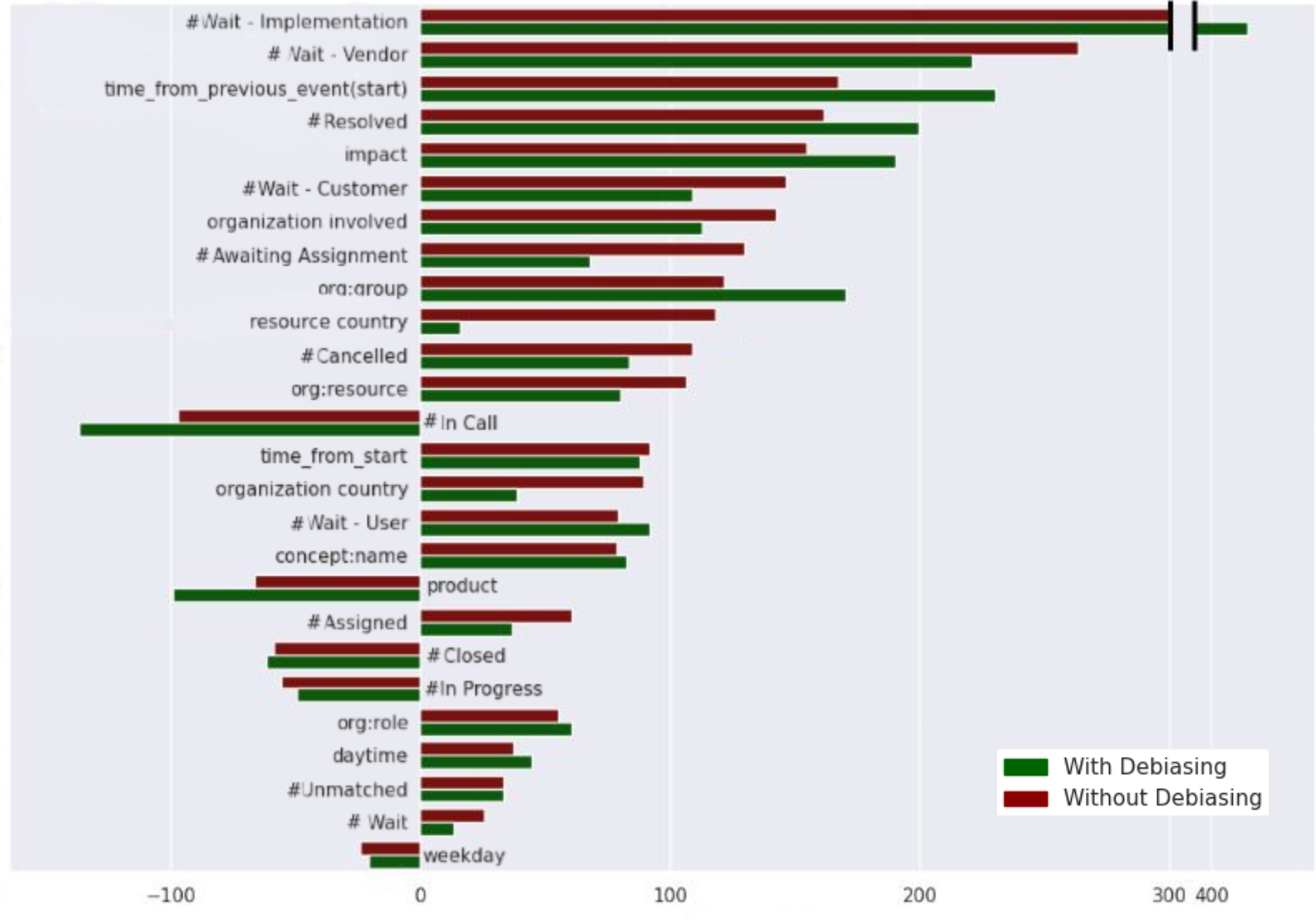}
    \caption{Shapley values for all variables for the VINST case study predicting the Total-Time outcome, sorted in descending order based on their absolute magnitudes. Shapley values are measured in hours.}
    \label{fig:VINST_time_shap}
\end{figure}

The assessment the effectiveness of our fairness framework to reduce the influence of the protected variables, we computed the Shapley values of the protected variables for the four case studies, both when we employed our framework and when we simply used the FCNN predictor that implements $\Phi$ (namely excluding the adversial FCNN for $\Phi_Z$). The results are reported in Table~\ref{tab:shap_drop}. In the case study related the VINST process for predicting the Total-Time outcome, the protected variable \textit{Resource country} is characterized by a Shapley value of 112 hours without using the debiasing framework, and 9 hours using the framework: the use of our framework brought the Shapley value down to 8\% of the value without using our framework, which is a remarkable result, given that the Shapley values are directly correlated with the feature importance in the prediction. For the same process, when the outcome was whether or not activity \textit{Awaiting Assignment Occurrence} is predicted to eventually occur, the protected variable \textit{Organization country} was characterized by a Shapley value that dropped from 1.8 to 0.03, when the debiasing framework was employed: the Shapley value has become 1\% of the value without debiasing. Similar results can be observed in Table~\ref{tab:shap_drop} for the other case studies, yielding the conclusion that \textit{observing the significant drop of the Shapley values of the protected value after applying our debiasing framework, the framework is extremely effective to reduce the influence of the protected variables and, thus, enhance the prediction fairness}.

\begin{figure}[t!]
    \centering
    \includegraphics[width=\linewidth]{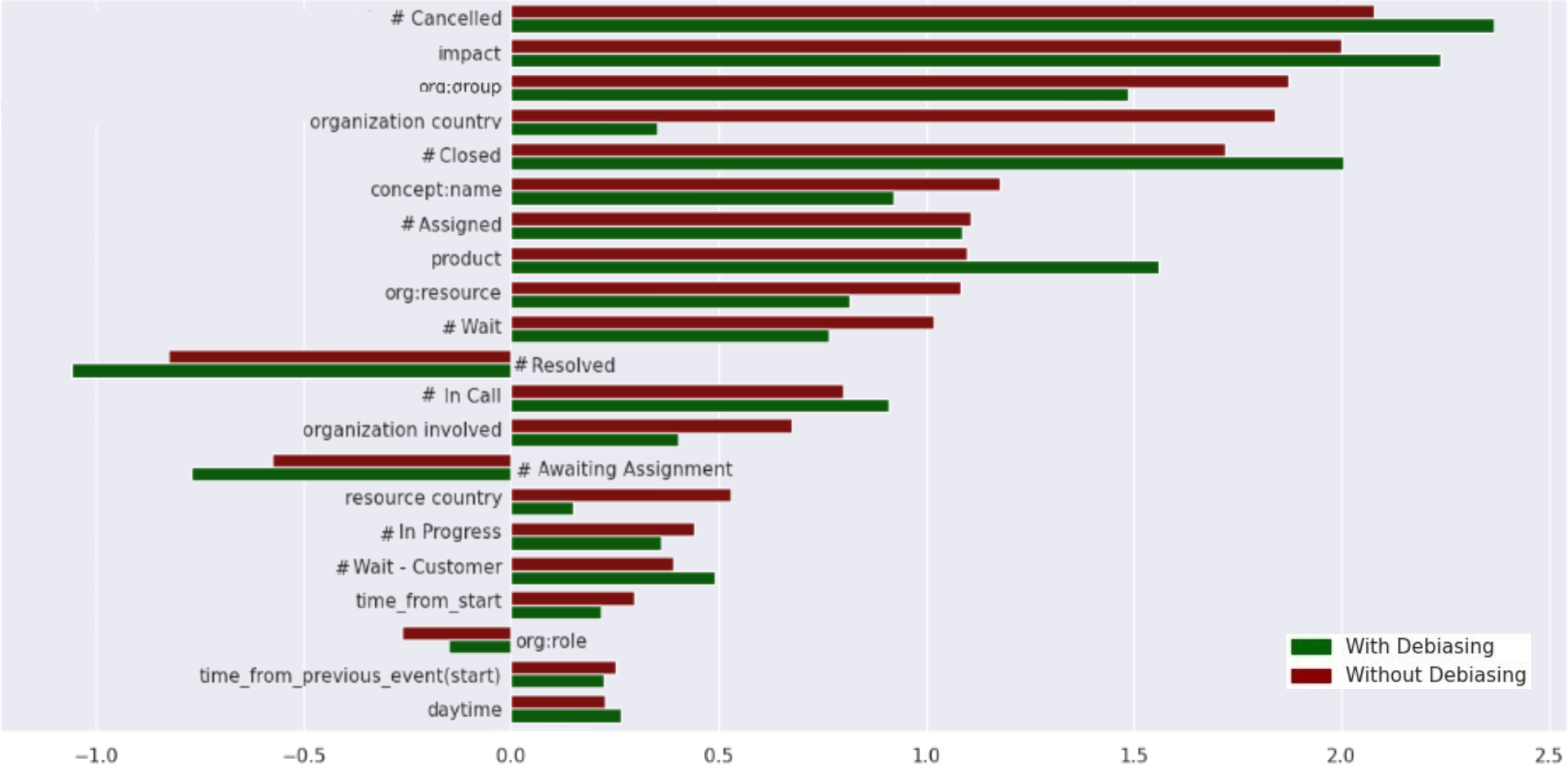}
    \caption{Shapley values for all variables for the VINST case study predicting the eventual occurrence of activity \textit{Awaiting Assignment}, sorted in descending order based on their absolute magnitudes.}
    \label{fig:VINST_act_shap}
\end{figure}

\begin{figure}[t!]
    \centering
    \includegraphics[width=\linewidth]{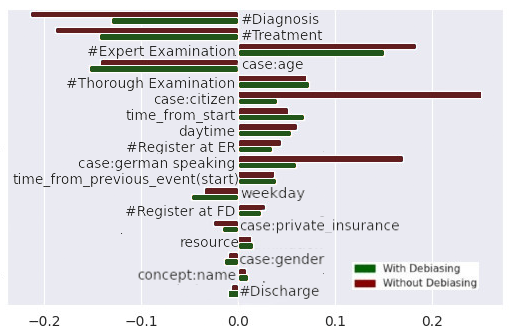}
    \caption{Shapley values for all variables for the hospital case study predicting the eventual occurrence of activity \textit{Treatment Unsuccessful}, sorted in descending order based on their absolute magnitudes.}
    \label{fig:Hospital_act_shap}
\end{figure}

\begin{table*}[t!]
\centering
\scriptsize
    \resizebox{\columnwidth}{!}
{
\begin{tabular}{|c|c|c|c|c|c|c|c|c|c|c|c|c|c|c|c|} 
\hline
& & \multicolumn{2}{|c|}{\textbf{Poland}} & \multicolumn{2}{|c|}{\textbf{Sweden}} &  \multicolumn{2}{|c|}{\textbf{India}}& \multicolumn{2}{|c|}{\textbf{Brazil}}&  \multicolumn{2}{|c|}{\textbf{Usa}} &  \multicolumn{2}{|c|}{\textbf{Std}} \\ \hline 
& & Without & With & Without & With  & Without & With  & Without & With  &  Without &With &  Without &With  \\
\hline 
\multirow{2}{*}{Qafari et al.~\cite{10.1007/978-3-030-33246-4_11}} & FPR & 0.20 & 0.18 & 0.13 & 0.24  & 0,11& 0,12& 0,17& 0,17&  0,32&0,41 &  0.143 & 0.086 \\ 
& TPR & 0.91 & 0.85 & 0.78 & 0.89 & 0.79 & 0.89 & 0.98 & 0.81 & 0.89 & 0.83 &  0.0641 & 0.0451 \\
\hline 
\multirow{2}{*}{Our framework} & FPR & 0.04 & 0.08 & 0.11 & 0.09  & 0,14& 0,08& 0,02& 0,06&  0,01&0,06 & 0.153 & 0.018 \\
& TPR & 0.67 & 0.61 & 0.72 & 0.63 & 0.62 & 0.63 & 0.59 & 0.65 & 0.59 & 0.65 & 0.052 & 0.024 \\
\hline
\end{tabular}
}
\caption{False Positive Rate (FPR) and True Positive Rate (TPR) achieved by the debiasing framework proposed here and by the framework by Qafari et al for the VINST process, aiming to predict the eventual occurrence of \textit{Awaiting Assignment}. Values are shown for the different groups, split by the values of the protected variables, together with the standard deviation of the FPR and TPR values for the different groups.}
\label{tab:rates_vinst}
\end{table*}
\begin{tabular}{c}\\
\end{tabular}

\begin{table*}[t]
    \scriptsize
    \centering
        \resizebox{\columnwidth}{!}
{
    \begin{tabular}{|c|c|c|c|c|c|c|c|c|c|c|c|c|c|} \hline 
         &  &  \multicolumn{6}{|c|}{\textbf{Citizen}}& \multicolumn{6}{|c|}{\textbf{german\_speaking}    \textbf{}}\\ \hline 
         &  &  \multicolumn{2}{|c|}{\textbf{True}}& \multicolumn{2}{|c|}{\textbf{False}}& \multicolumn{2}{|c|}{\textbf{Std}}& \multicolumn{2}{|c|}{\textbf{True}}&  \multicolumn{2}{|c|}{\textbf{False}}& \multicolumn{2}{|c|}{\textbf{Std}}\\ \hline 
         &  &  without&  with&   without& with& without& with&without&  with&  without&  with&  without& with \\\hline 
         \multirow{2}{*}{Qafari et al.~\cite{10.1007/978-3-030-33246-4_11}} &  FPR&  0.30 &  0.35&   0.36& 0.41& 0.03& 0.03&0.31&  0.35&  0.38&  0.40&  0.035& 0.025\\     
         &  TPR&  0.71&  0.67&   0.62& 0.51& 0.05& 0.08&0.71&  0.67&  0.62&  0.51&  0.09& 0.16\\ \hline 
         \multirow{2}{*}{Our framework} &  FPR&  0.28&  0.26&   0.22& 0.21& 0.03& 0.025&0.3&  0.24&  0.22&  0.21&  0.04& 0.015\\ 
         &  TPR&  0.82&  0.76&   0.77& 0.74& 0.025& 0.01&0.82&  0.76&  0.77&  0.74&  0.025& 0.01\\ \hline 
    \end{tabular}
    }
    \caption{False Positive Rate (FPR) and True Positive Rate (TPR) achieved by the debiasing framework proposed here and by the framework by Qafari et al for the Hospital process, aiming to predict the eventual occurrence of \textit{Treatment Unsuccessful}. Values are shown for the two groups, namely whether the patient is German or not (variable \textit{Citizen}) or whether the patient does or doesn't speak German (variable \textit{german\_speaking}). The standard deviation of the FPR and TPR between the two groups is also shown.}
    \label{tab:rates_pohl}
\end{table*}

Figure~\ref{fig:VINST_time_shap} shows all Shapley values for the VINST case study when the target outcome is the total time of instance executions: here, one can see that, indeed, the Shapley value for the protect variable \textit{resource country} has significantly dropped (cf.\ the green bar - when the debiasing framework is used - with the red bar - when it is not). One could also observe that the Shapley value for variable \textit{organization country} is also significantly reduced, likely because it is correlated with the protected variable. \textit{If we had simply removed the protected variable, the correlated variable \emph{organization country} would have gained strong influence onto the predictions: the bias would have simply moved from one sensitive variable to another, leaving the prediction model unfair. Conversely, our debiasing framework can also reduce the influence of the unfair variables that are strongly correlated to the one that has explicitly been stated as protected}. We can also observe that the Shapley value of other variables, such as \textit{time\_from\_previous\_event(start)} or \textit{\#Wait-Implementation}\footnote{Variable \textit{time\_from\_previous\_event(start)} models that time between the last event in the trace and the previous, while all variables starting with the hash symbol (\#) refer to the number of occurences of events for the specified activity.}, have increased their value: after debiasing, the prediction now leverages more on them. The Shapley values of all variables for the second case study for the VINST process is shown in Figure~\ref{fig:VINST_act_shap}, and one can see the same pattern as for the first, with reduction of the Shapley value for the protected variable \textit{organization country} and those correlated to it, along with the increase of other variables that are not discriminatory but become useful once the discriminatory can no longer be leveraged on.  

Space limitation only allows us to show the whole list of Shapley values for one more case study: we opted for the hospital case. The Shapley values are shown in Figure~\ref{fig:Hospital_act_shap}: along with the already-commented reduction of the value for the protected variables \textit{citizen} and \textit{german\_speaking}, a reduction can also be observed for the Shapley value for the variable \textit{private\_insurance}, which indicates whether or not the patient has a private complementary insurance: there is indeed a correlation between speaking German and subscribing a complementary insurance.

We complete the section by reporting the results with respect to the criterion of Equalized Odds (cf.\ Section~\ref{subsec:shap_e_odds}). This is only applicable to classification problems, and hence we can only verify the criterion for the case study related to the VINST process and the Hospital using the activity-occurrence process' outcome. We verified the extent to which we meet the criterion with our framework, and compared it with the similar results from Qafari et al.~\cite{10.1007/978-3-030-33246-4_11}. 

For the VINST case study related to the occurrence of activity \textit{Awaiting Assignment}, we considered the groups related to top five organization countries, which cover 89\% of the instances in the test set (recall that the protected variable is \textit{organization country}): Sweden, Poland, India, Brazil and USA. False positive and negative rates are reported in Table~\ref{tab:rates_vinst}, without and with using the framework, both for our framework and for that of Qafari et al.~\cite{10.1007/978-3-030-33246-4_11}, for all five groups. The last two columns with header \textit{Std} summarizes the standard deviation for FPR and TPR: in case of perfectly meeting the \textbf{Equalized Odds} criterion, there would be no difference among the groups, and thus the standard deviation would be zero. For our framework, the introduction of the debiasing phase, the FPR's standard deviation within the five groups is characterized by a 88\% drop, moving from $0.153$ to $0.018$, whereas the TPR's standard deviation shows a 53\% drop (from $0.052$ to $0.024$). \textit{Using the fairness approach by Qafari et al.~\cite{10.1007/978-3-030-33246-4_11}, the FPR's and TPR's standard deviation} within the five groups show a drop of 53\% and 29\%, which \textit{is nearly half the drop that our debiasing framework achieves}.
We conducted the same analysis for the hospital case study, which is reported in Table~\ref{tab:rates_pohl}. FPRs and TPRs are computed for both protected variables. 
Also for this case study, our debiasing framework guarantees lower FPR's and TPR's standard deviations for both variables, although the reduction is more limited than what achieved for the VINST case study.
However, The framework by Qafari et al.~\cite{10.1007/978-3-030-33246-4_11} does not reduce the FPR's and TPR's standard deviations for any of the two variables, expect for the FPR for variable \textit{german\_speaking}. As a matter of fact, their framework increases the TPR's standard deviation for both of variables, certainly going against the criterion of Equalized Odds.

\begin{table}[t!]
    \centering
    \resizebox{\columnwidth}{!}
{
    \begin{tabular}{|c|c|c|c|c|c|} \hline 
 \multirow{2}{*}{\textbf{Process}}& \multirow{2}{*}{\textbf{Predicted Outcome}} & \multicolumn{2}{c|}{\textbf{LSTM}} & \multicolumn{2}{c|}{\textbf{FCNN}} \\
 & & Training time & Accuracy & Training time & Accuracy \\ 
 \hline 
 VINST& Total Time & 30h 21min &76.2\%&6h 42min &78.0\%\\ \hline 
 \multirow{2}{*}{VINST} & Occurrence of & \multirow{2}{*}{39h 33min} & \multirow{2}{*}{0.81}& \multirow{2}{*}{7h 14min} & \multirow{2}{*}{0.80}\\ 
 & \textit{Awaiting Assignment} & & & & \\
 \hline 
 Hiring& Total Time& 45h 22min &82.3\%&14h 35min &83.6\%\\ \hline 
 \multirow{2}{*}{Hospital}& Occurrence of  & \multirow{2}{*}{41h 12min}&\multirow{2}{*}{0.78}&\multirow{2}{*}{10h 40min} & \multirow{2}{*}{0.78}\\ 
  & \textit{Treatment Unsuccessful} & & & & \\
 \hline
    \end{tabular}}
\caption{Training times for each case study. In the first column is represented the log name, while in the second  the case study. In the last four columns are reported the training times and the associated accuracies both for training the model using an LSTM network and a FCNN network.}
\label{tab:lstm_vs_fcnn}    
\end{table}

\subsection{Analysis of LSTM and FCNN accuracy and training times}\label{subsec:lstm_times}
We conclude this section by motivating the choice of FCNNs in place of LSTM models for operationalizing our framework. Specifically, we compared the prediction quality of the FCNN prediction models that implement $\Psi_\MK$ (cf.\ Figure~\ref{fig:framework}) for the four case studies, with that of equivalent models that leverage on a LSTM-network models. The LSTM-network models were trained via an implementation similar to~\cite{LSTM_camargo,galanti2020explainable}. We also measured the model training time to be able to assess the time amount necessary to build the prediction models.  
The results are reported in Table~\ref{tab:lstm_vs_fcnn}: the model-training times are significantly lower for FCNNs, while the prediction quality is very similar in every case study. 
This ultimately led us to conclude that FCNNs were preferable.

\section{Conclusion}
\label{sec:conclusion}
Considerable research efforts have been directed towards 
predictive process analytics. 
Section~\ref{sec:related_works} has shown that the fairness problem has generally been overlooked in predictive process analytics. 
This means that predictions may potentially be discriminatory, unethical, and, e.g., targeting certain ethnics, nationalities and religions.

This paper puts forward a predictive framework that specializes those based on adversarial debiasing so as to allow sequences (i.e., traces) as input. 

Experiments were carried out on three processes and four case studies, and the results show that our debasing framework minimizes the influence of the protected variables onto the prediction. At the same time, we illustrates that the reduction of the prediction quality is limited and lower than what is achieved by an existing framework for fairness-preserving process predictive analytics by Qafari et al.~\cite{10.1007/978-3-030-33246-4_11}. 

We acknowledge that we have used a specific encoding in our FCNNs for event-log traces (cf.\ Section~\ref{sucsec:preprocessing}), but alternatives are possible~\cite{barbon2023trace}. While this encoding has enable FCNNs to outperform the state of the art (see previous paragraph), 
a potential direction of future work is to evaluate alternative encodings, which can lead to further improvements.
So do we plan to extend our debiasing framework and move it from predictive to prescriptive analytics, in order to provide fair recommendations. A nice side effect on adding fairness to recommendations is that one can configure the system to give recommendations that are not biased on process' resources: this should ensure a fair assignment of recommended activities that do not cause overload of certain resources with respect to others. 


%

\end{document}